\pdfoutput=1

\documentclass[11pt]{article}
\usepackage{tcolorbox}
\usepackage[final]{acl}

\usepackage{times}
\usepackage{latexsym}

\usepackage[T1]{fontenc}

\usepackage[utf8]{inputenc}

\usepackage{microtype}

\usepackage{inconsolata}

\usepackage{graphicx}

\usepackage{amsmath}
\usepackage{array}
\usepackage{booktabs}
\usepackage{float}
\usepackage{multirow}
\usepackage{subcaption}
\usepackage{tablefootnote}
\usepackage{tabularx}

\title{One Sentence, Two Embeddings: Contrastive~Learning~of~Explicit~and~Implicit~Semantic~Representations}

\author{
  Kohei Oda\textsuperscript{1}{\qquad}Po-Min Chuang\textsuperscript{2}{\qquad}Kiyoaki Shirai\textsuperscript{1}{\qquad}Natthawut Kertkeidkachorn\textsuperscript{1} \\
  $^{1}$Japan Advanced Institute of Science and Technology \\
  $^{2}$Toshiba Corporation \\
  \textsuperscript{1}\texttt{\{s2420017, kshirai, natt\}@jaist.ac.jp} \\
  \textsuperscript{2}\texttt{pomin.chuang.x51@mail.toshiba}
}

\begin{document}
\maketitle
\begin{abstract}
Sentence embedding methods have made remarkable progress, yet they still struggle to capture the implicit semantics within sentences. 
This can be attributed to the inherent limitations of conventional sentence embedding methods that assign only a single vector per sentence. 
To overcome this limitation, we propose DualCSE, a sentence embedding method that assigns two embeddings to each sentence: one representing the explicit semantics and the other representing the implicit semantics. 
These embeddings coexist in the shared space, enabling the selection of the desired semantics for specific purposes such as information retrieval and text classification. 
Experimental results demonstrate that DualCSE can effectively encode both explicit and implicit meanings and improve the performance of the downstream task.\footnote{Our code is publicly available at \url{https://github.com/iehok/DualCSE}.}
\end{abstract}

\section{Introduction}
Sentence embeddings have been extensively studied in the field of natural language processing \citep{reimers-gurevych-2019-sentence,jiang-etal-2022-promptbert,li2025ese}. 
However, most existing sentence embedding methods struggle to capture implicit semantics.\footnote{In this paper, the term ``explicit semantics'' is employed to denote literal meanings, while ``implicit semantics'' is used to indicate non-literal meanings derived from figurative or pragmatic usage.}
\citet{sun2025textembeddingscaptureimplicit} pointed out even state-of-the-art sentence embedding methods \citep{wang2024textembeddingsweaklysupervisedcontrastive,zhang-etal-2024-mgte,zhang2025jasperstelladistillationsota} exhibit a nearly 20\% performance gap between explicit and implicit semantics on the MTEB classification benchmark \citep{muennighoff-etal-2023-mteb}. 
This may be due to the limitation of existing methods, which assign only a single vector to a sentence and overlook the presence of multiple interpretations. 

\begin{figure}[t]
    \centering
    \includegraphics[width=1.0\linewidth]{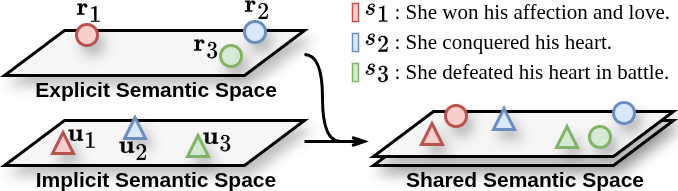}
    \caption{Overview of DualCSE. The explicit and implicit semantic spaces are combined into a shared space.}
    \label{fig:overview}
\end{figure}

To address this limitation, we propose DualCSE, a \underline{dual}-semantic \underline{c}ontrastive \underline{s}entence \underline{e}mbedding framework that assigns two embeddings to each sentence: one representing its explicit semantic and the other representing its implicit semantic. 
As shown in Figure~\ref{fig:overview}, the explicit and implicit semantics of sentences are represented in the shared space by DualCSE. 
For example, the explicit semantic of ``She conquered his heart.''($s_2$) is close to the explicit semantic of ``She defeated his heart in battle.''($s_3$), and the implicit semantic of $s_2$ is close to the explicit semantic of ``She won his affection and love.''($s_1$). 
Furthermore, for each of $s_1$ and $s_3$, the similarity between the explicit and implicit semantics is higher than the that of $s_2$. 
Our method not only provides useful features for fundamental tasks such as information retrieval \citep{thakur2021beir} and text classification \citep{maas-etal-2011-learning}, but also facilitates the estimation of the implicit nature of a given sentence \citep{wang2025impscore}. 
DualCSE is trained via contrastive learning \citep{pmlr-v119-chen20j} using natural language inference (NLI) datasets based on representative supervised sentence-embedding methods \citep{gao-etal-2021-simcse, ni-etal-2022-sentence, li-li-2024-aoe}. 
Specifically, we leverage an NLI dataset considering both explicit and implicit semantics \citep{havaldar-etal-2025-entailed} as training data and utilize a novel contrastive loss. 

To evaluate the capability of DualCSE in capturing inter-sentence and intra-sentence relations, we conduct two experiments of two tasks: Recognizing Textual Entailment (RTE) and Estimating Implicitness Score (EIS). 
Experimental results show that DualCSE captures inter- and intra-sentence relations more accurately than conventional methods.

\section{Implied NLI (INLI) Dataset}
The INLI dataset \citep{havaldar-etal-2025-entailed} is used for DualCSE. 
As shown in Table~\ref{tab:inli_example}, the INLI dataset differs from standard NLI datasets such as SNLI \citep{bowman-etal-2015-large} and MNLI \citep{williams-etal-2018-broad} in that it provides four different hypotheses, labeled with ``implied-entailment'', ``explicit-entailment'', ``neutral'', and ``contradiction'' for a single premise. 
The implied-entailment and explicit-entailment indicate entailment with respect to the implicit and explicit semantics of the premise, respectively.\footnote{The detailed statistics of the INLI dataset are shown in the Appendix~\ref{app:dataset_statistics}.}

\begin{table}[t]
    \centering
    \small
    \begin{tabular}{l}
        \toprule
        \textbf{Premise} \\
        Diane says, ``Would you like to go a party tonight?'' \\Sophie responds, ``I am too tired.'' \\
        \midrule
        \textbf{Implied Entailment} \\
        Sophie would prefer not to attend the party this evening. \\
        \midrule
        \textbf{Explicit Entailment} \\
        Sophie claims to be too tired. \\
        \midrule
        \textbf{Neutral} \\
        The party will take place outside. \\
        \midrule
        \textbf{Contradiction} \\
        Sophie is excited to attend the party this evening. \\
        \bottomrule
    \end{tabular}
    \caption{An example of a sample in the INLI dataset}
    \label{tab:inli_example}
\end{table}

\section{DualCSE}
This section presents DualCSE, a method that encodes each sentence $s$ into two embeddings: $\mathbf{r}$, representing its explicit semantics and $\mathbf{u}$, representing its implicit semantics. 
The loss function for learning these embeddings is first explained, followed by a description of the model architecture. 

\subsection{Contrastive Loss}
\begin{figure}[t]
    \centering
    \includegraphics[width=0.8\linewidth]{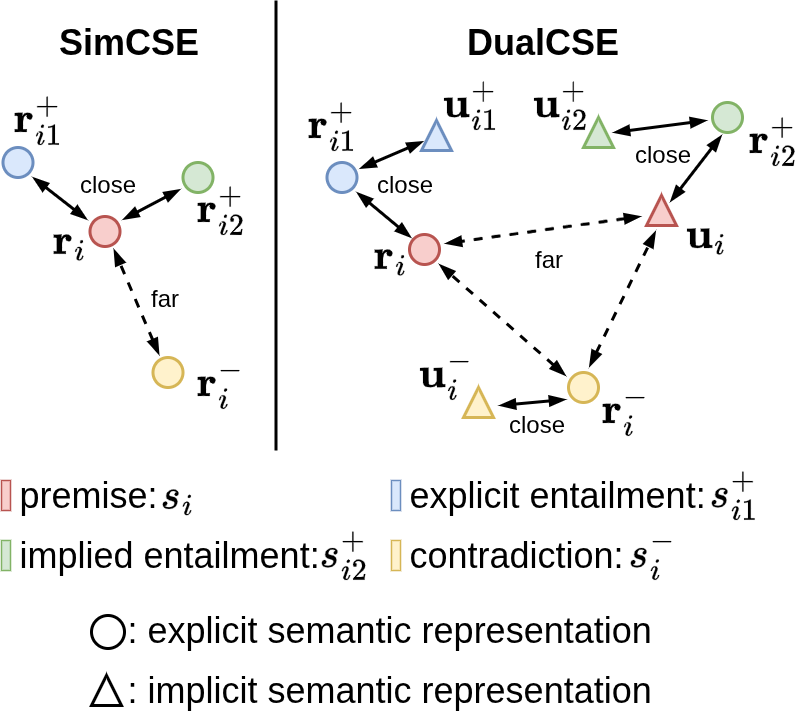}
    \caption{Conceptual diagram of contrastive loss in SimCSE \citep{gao-etal-2021-simcse} and our DualCSE.}
    \label{fig:loss}
\end{figure}

For a given sample in the INLI dataset, let $s_i$ be a premise, and $s^+_{i1}$, $s^+_{i2}$, and $s^-_i$ be the explicit-entailment, implied-entailment, and contradiction hypothesis for $s_i$, respectively. 
The explicit-semantic embeddings are denoted as $\mathbf{r}_i$, $\mathbf{r}^+_{i1}$, $\mathbf{r}^+_{i2}$, and $\mathbf{r}^-_i$, while the implicit-semantic embeddings are denoted as $\mathbf{u}_i$, $\mathbf{u}^+_{i1}$, $\mathbf{u}^+_{i2}$, and $\mathbf{u}^-_i$. 
The contrastive loss $l_i$ for $i$-th instance in a batch of size $N$ is calculated as follows: 
\begin{equation}
    v(\mathbf{h}_1, \mathbf{h}_2) = e^{\mathrm{sim}(\mathbf{h}_1, \mathbf{h}_2)/\tau}, 
\end{equation}
\vspace{-1em}
\begin{equation}
    \begin{array}[b]{@{}l@{}}
        l_i = \\
        \displaystyle
        -\log{\frac{v(\mathbf{r}_i, \mathbf{r}^+_{i1})}{\sum^{N}_{j=1}(v(\mathbf{r}_i, \mathbf{r}^+_{j1}) + v(\mathbf{r}_i, \mathbf{r}^-_j) + v(\mathbf{r}_i, \mathbf{u}_j))}} \\
        \displaystyle
        - \log{\frac{v(\mathbf{u}_i, \mathbf{r}^+_{i2})}{\sum^{N}_{j=1}(v(\mathbf{u}_i, \mathbf{r}^+_{j2}) + v(\mathbf{u}_i, \mathbf{r}^-_j) + v(\mathbf{u}_i, \mathbf{r}_j))}} \\
        \displaystyle
        - \log{\frac{v(\mathbf{r}^+_{i1}, \mathbf{u}^+_{i1})}{\sum^{N}_{j=1}v(\mathbf{r}^+_{i1}, \mathbf{u}^+_{j1})}} - \log{\frac{v(\mathbf{r}^+_{i2}, \mathbf{u}^+_{i2})}{\sum^{N}_{j=1}v(\mathbf{r}^+_{i2}, \mathbf{u}^+_{j2})}} \\
        \displaystyle
        - \log{\frac{v(\mathbf{r}^-_i, \mathbf{u}^-_{i})}{\sum^{N}_{j=1}v(\mathbf{r}^-_i, \mathbf{u}^-_j)}}, 
    \end{array}
    \label{eq:loss}
\end{equation}
where $\mathrm{sim}(\mathbf{h}_1, \mathbf{h}_2)$ denotes the cosine similarity between $\mathbf{h}_1$ and $\mathbf{h}_2$, and $\tau$ is the temperature parameter. 
Intuitively, as shown in Figure~\ref{fig:loss}, the pairs ($\mathbf{r}_i$, $\mathbf{r}^+_{i1}$) and ($\mathbf{u}_i$, $\mathbf{r}^+_{i2}$) are encouraged to close together, whereas the pairs ($\mathbf{r}_i$, $\mathbf{r}^-_i$) and ($\mathbf{u}_i$, $\mathbf{r}^-_i$) are encouraged to push apart.\footnote{This is encoded in the first and second terms on the right-hand side of Equation~(\ref{eq:loss}).}
These are designed to capture inter-sentence relations, i.e., a premise and entailment hypothesis are similar, while a premise and contradiction hypothesis are dissimilar. 
Furthermore, the pairs ($\mathbf{r}^+_{i1}$, $\mathbf{u}^+_{i1}$), ($\mathbf{r}^+_{i2}$, $\mathbf{u}^+_{i2}$), and ($\mathbf{r}^-_i$, $\mathbf{u}^-_i$) are encouraged to close together,\footnote{This is encoded in the third, fourth, and fifth terms in Equation~(\ref{eq:loss}).} whereas the pair ($\mathbf{r}_i$, $\mathbf{u}_i$) is encouraged to push apart.\footnote{This is encoded in $v(\mathbf{r}_i, \mathbf{u}_j)$ in the denominator of the first term and $v(\mathbf{u}_i, \mathbf{r}_j)$ in the second term in Equation~(\ref{eq:loss}).}
These are designed to capture intra-sentence relations under the assumption that the hypotheses in the INLI dataset are less ambiguous and convey more similar explicit and implicit semantics than a premise. 

\subsection{Model Architecture}
This study employs two types of encoder models as follows. 
\paragraph{Cross-encoder}
A single BERT \citep{devlin-etal-2019-bert} or RoBERTa \citep{liu2019robertarobustlyoptimizedbert} model that outputs the representation $\mathbf{r}$ for the explicit semantic of $s$ when given the input ``\texttt{[CLS]} $s$ \texttt{[SEP]} explicit,'' and $\mathbf{u}$ for the implicit semantic of $s$ when given the input ``\texttt{[CLS]} $s$ \texttt{[SEP]} implicit.''
\paragraph{Bi-encoder}
Two separate BERT or RoBERTa models are trained to obtain $\mathbf{r}$ and $\mathbf{u}$, respectively. 

For both models, the hidden state of the final layer of \texttt{[CLS]} is used as the sentence embedding.

\section{Experiments}
We validate the effectiveness of DualCSE through experiments on two tasks. 
The first task is Recognizing Textual Entailment (RTE), which involves the model’s capacity to correctly capture entailment relationships between sentences. 
The second task is Estimating Implicitness Score (EIS), which aims to estimate the extent to which an implicit meaning deviates from a literal meaning. 

\subsection{Experimental Setup}
For the two model architectures of DualCSE, the pre-trained BERT\textsubscript{base} and RoBERTa\textsubscript{base} are employed as the encoder models. 
Only the settings and results of the RoBERTa model are reported in this section, since it demonstrated higher performance than BERT on the development set. 
The batch size and learning rate are optimized using the development set, resulting in 64 and 5e-5 for the cross-encoder, and 32 and 3e-5 for the bi-encoder.\footnote{The optimization results are shown in Appendix \ref{app:hyperparameter_optimization}.}
The temperature parameter $\tau$ is set to 0.05, following \citet{gao-etal-2021-simcse} and \citet{yoda-etal-2024-sentence}. 

\subsection{Recognizing Textual Entailment (RTE)}
\paragraph{Task definition}
RTE is a task that classifies a given premise and hypothesis pair ($p$, $h$) as either ``entailment'' or ``non-entailment.'' 
The INLI dataset \citep{havaldar-etal-2025-entailed} is used for the experiment, where the neutral and contradiction labels are converted to ``non-entailment,'' and both explicit and implied entailment are retained as ``entailment.''

\paragraph{Method}
Let $\mathbf{r}_1$ and $\mathbf{r}_2$ be the representations of the explicit semantics of the premise $p$ and hypothesis $h$, respectively, and $\mathbf{u}_1$ be the representation of the implicit semantics of $p$. 
DualCSE predicts that $p$ and $h$ are in an entailment relation if 
\begin{equation}
\max\bigl(\cos(\mathbf{r}_1, \mathbf{r}_2), \cos(\mathbf{u}_1, \mathbf{r}_2)\bigr) > \gamma, 
\end{equation}
and predicts non-entailment otherwise. 
The threshold $\gamma$ is tuned on the INLI development set. 

\paragraph{Baselines}
Two baselines are compared to DualCSE: SimCSE (SNLI+MNLI) \citep{gao-etal-2021-simcse} and SimCSE (INLI). 
The latter is a SimCSE model trained on the INLI dataset. 
These baselines predict labels using the same approach as our model, which involves determining whether the cosine similarity between the premise and hypothesis embeddings exceeds the threshold. 
Additionally, for reference, we also provide the results of a few-shot setting with large language models (LLMs).\footnote{Detailed prompts are provided in Appendix~\ref{app:prompt}.}

\begin{table}[t]
\centering
\small
\setlength{\tabcolsep}{2pt}
\begin{tabular}{lccccc}
\toprule
\textbf{Model} & \textbf{Exp.} & \textbf{Imp.} & \textbf{Neu.} & \textbf{Con.} & \textbf{Avg.} \\
\midrule
SimCSE (SNLI+MNLI)      & 79.80 & 49.00 & 74.30 & 67.60 & 67.68 \\
SimCSE (INLI)           & 90.60 & 69.10 & 66.90 & 91.00 & 79.40 \\
DualCSE-Cross (ours)    & 90.20 & 73.40 & 68.40 & 88.70 & 80.18 \\
DualCSE-Bi (ours)       & 91.90 & 69.90 & 72.10 & 87.60 & \textbf{80.38} \\
\midrule
Gemini-1.5-Pro          & 97.90 & 80.30 & 92.00 & 95.40 & 91.40 \\
\bottomrule
\end{tabular}
\caption{Results of RTE task (accuracy \%). \textbf{Exp.}, \textbf{Imp.}, \textbf{Neu.}, and \textbf{Con.} mean the accuracy for the instances where the original label is explicit-entailment, implied-entailment, neutral, and contradiction, respectively.}
\label{tab:rte-results}
\end{table}

\begin{table*}[t]
    \centering
    \small
    \begin{tabular}{l|l}
        \toprule
        \multicolumn{2}{l}{\textbf{Query}: Madeleine has just moved into a neighbourhood and meets her new neighbour Pierre.} \\
        \multicolumn{2}{l}{Pierre says, ``Are you from this state?'' Madeleine responds, ``I'm from Oregon.''} \\
        \midrule
        \textbf{Explicit semantic}: Madeleine is from Oregon. & \textbf{Implicit semantic}: Madeleine was born in a different state. \\
        \midrule
        \#1 Laverne moved from Canada. & \#1 The place does not belong to Quincy. \\
        \#2 Angela and her family live in Portland now. & \#2 Madeleine enjoys food with some spice, but not if it's overly hot. \\
        \#3 Alyce works in Portland. & \#3 Earlene is not originally from this area. \\
        \bottomrule
    \end{tabular}
    \caption{An example of a simple retrieval experiment. \textbf{Explicit semantic} and \textbf{Implicit semantic} are the explicit-entailment and implied-entailment hypotheses in the INLI dataset, respectively. These are not used as a query for sentence retrieval.}
    \label{tab:retrieval_example}
\end{table*}

\paragraph{Results}
The results are shown in Table~\ref{tab:rte-results}. 
First, the proposed method DualCSE outperforms SimCSE (INLI) in both model architectures, demonstrating the effectiveness of representations for the explicit and implicit semantics of sentences. 
Comparing the cross-encoder and bi-encoder of DualCSE, the cross-encoder is superior for the implied-entailment samples, while the bi-encoder is better for the neutral ones. 
However, the overall performance is almost identical. 
Next, SimCSE (SNLI+MNLI) has the largest gap in accuracy between Exp. and Imp. 
This is likely due to SNLI and MNLI containing relatively few sentences with implicit semantics, as reported by \citet{havaldar-etal-2025-entailed}. 
Finally, LLMs generally demonstrate superior performance compared to the encoder models. 
However, similar to other models, LLMs consistently show a tendency toward lower performance on Imp. compared to Exp.\footnote{The results of other LLMs are provided in Appendix \ref{app:full_results_of_rte}.}

\subsection{Estimating Implicitness Score (EIS)}
\paragraph{Task definition}
Given two sentences $s_1$ and $s_2$, predict which sentence exhibits a higher degree of implicitness. 
Two datasets are employed for this task: the INLI \citep{havaldar-etal-2025-entailed} and the dataset provided by \citet{wang2025impscore}. 
For the INLI, it is supposed that the premise is more implicit than the hypothesis. 

\paragraph{Method}
The implicitness score of a sentence $s$ is calculated as follows: 
\begin{equation}
    \mathrm{imp}(s) = 1 - \cos(\mathbf{r}, \mathbf{u}). 
\end{equation}
We predict which of the sentences $s_1$ and $s_2$ has the greater implicitness score: 
\begin{equation}
\arg\max\bigl(\mathrm{imp}(s_1), \mathrm{imp}(s_2)\bigr). 
\end{equation}

\paragraph{Baselines}
Three baselines are compared in this experiment: (1) \textsc{Length}, which chooses the longer sentence, (2) ImpScore (original) \citep{wang2025impscore}, and (3) ImpScore (INLI), which is the ImpScore trained on the INLI dataset using RoBERTa as the encoder model. 

\begin{table}[t]
\centering
\small
\begin{tabular}{lcc}
\toprule
\textbf{Model}          & \textbf{INLI} & \textbf{\citet{wang2025impscore}} \\
\midrule
\textsc{Length}         & 99.90         & 73.37 \\
ImpScore (original)     & 80.55         & \textbf{95.20} \\
ImpScore (INLI)         & 99.97         & 81.56 \\
DualCSE-Cross (ours)    & 99.97         & 79.31 \\
DualCSE-Bi (ours)       & \textbf{100}  & 77.48 \\
\bottomrule
\end{tabular}
\caption{ESI task results (accuracy \%)}
\label{tab:eis_results}
\end{table}

\paragraph{Results}
The results are shown in Table~\ref{tab:eis_results}. 
First, DualCSE achieves near-perfect accuracy in both model architectures for the INLI dataset, i.e., for the in-domain setting. 
However, this may be because the length ratio of the input sentence pairs serves as a useful signal, as evidenced by the near-perfect performance accuracy achieved by \textsc{Length} as well. 
Next, in the out-of-domain setting (Wang's dataset), the accuracy of DualCSE and ImpScore (INLI) decreases to nearly 80\%.
The performance of DualCSE is comparable to that of ImpScore. 
It is worth noting that the ImpScore has been developed specifically for the purpose of predicting the implicitness score, whereas our DualCSE is capable of generating embeddings for both explicit and implicit semantics, which enables it to perform other downstream tasks. 
In addition, consistent with the RTE task, the performance of the cross-encoder and bi-encoder of DualCSE is comparable.\footnote{The results of other models are provided in Appendix \ref{app:full_results_of_eis}.}

\section{Analysis}

\begin{table}[t]
    \centering
    \small
    \begin{tabular}{lcc}
        \toprule
        \textbf{Loss function}                  & \textbf{RTE}      & \textbf{EIS} \\
        \midrule
        DualCSE-Cross                           & \textbf{80.18}    & \textbf{99.97} \\
        ~~w/o contradiction                     & 64.57             & 99.88 \\
        ~~w/o intra sentence                    & 80.10             & 92.25 \\
        ~~w/o contradiction \& intra sentence   & 64.68             & 32.75 \\
        \bottomrule
    \end{tabular}
    \caption{Ablation results (accuracy \%)}
    \label{tab:ablation_results}
\end{table}

\paragraph{Ablation Study}
The ablation study is conducted to investigate the contributions of the components of the proposed contrastive loss. 
We train the models in three scenarios: excluding the loss for contradiction hypotheses, excluding the loss for intra-sentence relations, and excluding both. 
As shown in Table~\ref{tab:ablation_results}, the loss for contradiction hypotheses is more effective for the RTE task, while the loss for intra-sentence relations is more effective for the EIS task.\footnote{More details of the ablation and the results of other models are described in Appendix \ref{app:ablation_results}.}

\paragraph{Retrieval Experiment}
A qualitative evaluation of the explicit and implicit embeddings is conducted through a simple search experiment. 
Specifically, we select several premises from the development data of INLI as queries and retrieve the top three similar hypotheses from the training data that most closely match the explicit and implicit semantics of each query. 
As shown in Table~\ref{tab:retrieval_example}, DualCSE facilitates a separate search for the sentences that correspond to explicit and implicit semantics.\footnote{Other examples are described in Appendix \ref{app:retrieval_examples}.}

\section{Conclusion}
This paper proposed DualCSE, a sentence embedding method that assigns two representations for the explicit and implicit semantics of sentences. 
The experimental results of the RTE and EIS tasks demonstrated DualCSE successfully encoded literal and latent meanings into separate embeddings.

\section*{Limitations}
We use only the INLI \citep{havaldar-etal-2025-entailed} dataset for the training. 
However, the variation of the sentences in the INLI is rather limited. 
It is important to apply DualCSE to the training data of various domains. 
For example, the datasets for hate speech detection \citep{hartvigsen-etal-2022-toxigen} and sentiment analysis \citep{pontiki-etal-2014-semeval} are converted to the INLI format and can be used as the training data. 

This study conducted a simple retrieval experiment with the aim of applying it to real-world applications. 
In the future, it would be desirable to apply our method to more practical settings, such as analyzing customer reviews and implementing search engines. 

Recently, sentence embedding methods using LLMs have been actively studied \citep{behnamghader2024llmvec,jiang-etal-2024-scaling,yamada2025outoftheboxconditionaltextembeddings}. 
Extending our method to LLMs is another future direction. 

Finally, since the meaning of ``implicit semantics'' in this study is largely derived from the implicit-entailment of the INLI dataset, further investigation is required to determine whether they can be generalized to broader linguistic pragmatics.

\bibliography{anthology1,anthology2,custom}

\appendix

\section{Dataset Statistics}
\label{app:dataset_statistics}
Table~\ref{tab:dataset_detail} shows the statistics of the INLI dataset and Wang's dataset \citep{wang2025impscore}. 

\begin{table}[!ht]
\small
\centering
\begin{tabular}{lccc}
\toprule
\textbf{Dataset}            & \textbf{train}    & \textbf{development}  & \textbf{test} \\
\midrule
INLI                        & ~~32,000          & 4,000                 & 4,000 \\
\citet{wang2025impscore}    & 101,320           & 5,630                 & 5,630 \\
\bottomrule
\end{tabular}
\caption{Statistics of datasets. This table shows the number of premise-hypothesis pairs for the INLI dataset and that of implicit-explicit sentence pairs for Wang’s dataset.}
\label{tab:dataset_detail}
\end{table}

\section{Hyperparameter Optimization}
\label{app:hyperparameter_optimization}
The hyperparameters, i.e., batch size and learning rate, are optimized via a grid search. 
The results of the grid search are shown in Table~\ref{tab:hypara_opt}. 
The time and GPU memory required for training are shown in Table~\ref{tab:time_and_memory}. 
All experiments were conducted on a 20 GB NVIDIA H100 MIG instance (a quarter of a full H100). 

\begin{table}[!ht]
    \centering
    \small
    \begin{subtable}{1.0\linewidth}
    \centering
    \begin{tabular}{lccc}
        \toprule
        \multirow{2}{*}{Batch size} & \multicolumn{3}{c}{Learning rate} \\
                                    & 1e-5  & 3e-5  & 5e-5 \\
        \midrule
                                16  & 76.50 & 77.10 & 77.53 \\
                                32  & 76.40 & 77.42 & \textbf{77.70} \\
                                64  & 75.42 & 76.92 & 77.45 \\
        \bottomrule
    \end{tabular}
    \caption{DualCSE-Cross-BERT}
    \label{tab:hypara_opt_cross_bert}
    \end{subtable}
    
    \bigskip
    
    \begin{subtable}{1.0\linewidth}
    \centering
    \begin{tabular}{lccc}
        \toprule
        \multirow{2}{*}{Batch size} & \multicolumn{3}{c}{Learning rate} \\
                                    & 1e-5  & 3e-5  & 5e-5 \\
        \midrule
                                16  & 76.92 & \textbf{78.57} & 78.30 \\
                                32  & 76.23 & 78.07 & 78.47 \\
                                64  & 75.45 & 77.25 & 77.97 \\
        \bottomrule
    \end{tabular}
    \caption{DualCSE-Bi-BERT}
    \label{tab:hypara_opt_bi_bert}
    \end{subtable}
    
    \bigskip
    
    \begin{subtable}{1.0\linewidth}
    \centering
    \begin{tabular}{lccc}
        \toprule
        \multirow{2}{*}{Batch size} & \multicolumn{3}{c}{Learning rate} \\
                                    & 1e-5  & 3e-5  & 5e-5 \\
        \midrule
                                16  & 80.50 & 80.60 & 80.58 \\
                                32  & 79.85 & 80.75 & 81.12 \\
                                64  & 78.83 & 79.93 & \textbf{81.15} \\
        \bottomrule
    \end{tabular}
    \caption{DualCSE-Cross-RoBERTa}
    \label{tab:hypara_opt_cross_roberta}
    \end{subtable}
    
    \bigskip
    
    \begin{subtable}{1.0\linewidth}
    \centering
    \begin{tabular}{lccc}
        \toprule
        \multirow{2}{*}{Batch size} & \multicolumn{3}{c}{Learning rate} \\
                                    & 1e-5  & 3e-5  & 5e-5 \\
        \midrule
                                16  & 80.40 & 80.45 & 80.33 \\
                                32  & 80.60 & \textbf{80.80} & 80.45 \\
                                64  & 79.47 & 80.55 & 80.65 \\
        \bottomrule
    \end{tabular}
    \caption{DualCSE-Bi-RoBERTa}
    \label{tab:hypara_opt_bi_roberta}
    \end{subtable}
    
    \bigskip
    
    \begin{subtable}{1.0\linewidth}
    \centering
    \begin{tabular}{lccc}
        \toprule
        \multirow{2}{*}{Batch size} & \multicolumn{3}{c}{Learning rate} \\
                                    & 1e-5  & 3e-5  & 5e-5 \\
        \midrule
                                16  & 76.38 & 76.82 & 76.63 \\
                                32  & 76.65 & \textbf{77.35} & 77.25 \\
                                64  & 75.65 & 76.40 & 76.40 \\
        \bottomrule
    \end{tabular}
    \caption{SimCSE-BERT}
    \label{tab:hypara_opt_simcse_bert}
    \end{subtable}
    
    \bigskip
    
    \begin{subtable}{1.0\linewidth}
    \centering
    \begin{tabular}{lccc}
        \toprule
        \multirow{2}{*}{Batch size} & \multicolumn{3}{c}{Learning rate} \\
                                    & 1e-5  & 3e-5  & 5e-5 \\
        \midrule
                                16  & 80.12 & 79.18 & 78.30 \\
                                32  & 79.52 & \textbf{80.43} & 79.72 \\
                                64  & 79.43 & 79.85 & 79.37 \\
        \bottomrule
    \end{tabular}
    \caption{SimCSE-RoBERTa}
    \label{tab:hypara_opt_simcse_roberta}
    \end{subtable}
    \caption{Grid search results}
    \label{tab:hypara_opt}
\end{table}

\begin{table}[!ht]
    \centering
    \small
    \setlength{\tabcolsep}{3pt}
    \begin{tabular}{lccc}
        \toprule
        \multirow{2}{*}{Model}  & \multicolumn{3}{c}{Batch size} \\
                                & 16            & 32            & 64 \\
        \midrule
        SimCSE                  & 24.26 / 4.29  & 12.52 / 5.09  & 7.01 / 7.05 \\
        DualCSE-Cross           & 31.76 / 6.51  & 17.27 / 8.56  & 9.76 / 13.73 \\
        DualCSE-Bi              & 31.01 / 8.39  & 16.76 / 10.55 & 9.01 / 15.42 \\
        \bottomrule
    \end{tabular}
    \caption{The time (minutes) / GPU memory (GB) required for training}
    \label{tab:time_and_memory}
\end{table}

\section{Prompt}
\label{app:prompt}
The prompt used for the RTE task is shown in Figure~\ref{fig:prompt}, which is shared across the following LLMs: GPT-4, GPT-4o, GPT-4o-mini, Claude-3.7-Sonnet, Gemini-1.5-Pro, Gemini-2.0-Flash, DeepSeek-v3 and Mistral-Large. 
For each test data point, eight sentence pairs are randomly selected from the training data and included in the prompt for few-shot learning, with an equal number of entailment and non-entailment pairs.

\begin{figure}[!ht]
\centering
\small
\begin{tcolorbox}[colback=white, colframe=black, boxrule=0.5pt, arc=2pt, left=3pt, right=3pt, top=3pt, bottom=3pt]
\begin{flushleft}

\texttt{Given a premise and a hypothesis, your task is to label whether the hypothesis is a valid inference from the premise.}\\
\texttt{Specifically, you will need to assign one of two labels to the hypothesis:}

\vspace{0.5em}
\texttt{Label: entailment}\\
\texttt{Definition: The hypothesis is a valid inference from the passage, but it is NOT explicitly stated in the passage or the hypothesis is a valid inference from the passage, and it is explicitly stated in the passage.}

\vspace{0.5em}
\texttt{Label: non\_entailment}\\
\texttt{Definition: Not including Implicature or Explicature entailment.}

\vspace{0.5em}
\texttt{[----- begin examples -----]}\\
\texttt{Premise: \{premise\} Hypothesis: \{hypothesis\} Label: entailment}\\
\texttt{Premise: \{premise\} Hypothesis: \{hypothesis\} Label: entailment}\\
\texttt{Premise: \{premise\} Hypothesis: \{hypothesis\} Label: entailment}\\
\texttt{Premise: \{premise\} Hypothesis: \{hypothesis\} Label: entailment}\\
\texttt{Premise: \{premise\} Hypothesis: \{hypothesis\} Label: non\_entailment}\\
\texttt{Premise: \{premise\} Hypothesis: \{hypothesis\} Label: non\_entailment}\\
\texttt{Premise: \{premise\} Hypothesis: \{hypothesis\} Label: non\_entailment}\\
\texttt{Premise: \{premise\} Hypothesis: \{hypothesis\} Label: non\_entailment}\\
\texttt{[----- end examples -----]}

\vspace{0.5em}
\texttt{[Your Task] Given a premise and a hypothesis, your task is to label the hypothesis as one of the two labels: entailment, non\_entailment.}\\
\texttt{Your response should be only one word, the name of the label.}\\
\texttt{Premise: \{premise\} Hypothesis: \{hypothesis\} Label:}
\end{flushleft}
\end{tcolorbox}
\caption{Prompt template for RTE task}
\label{fig:prompt}
\end{figure}

\section{Full Results of RTE}
\label{app:full_results_of_rte}
The full results of the RTE task are shown in Table~\ref{tab:rte_results_full}. 

\begin{table}[!ht]
\centering
\small
\setlength{\tabcolsep}{2pt}
\begin{tabular}{lccccc}
\toprule
\textbf{Model} & \textbf{Exp.} & \textbf{Imp.} & \textbf{Neu.} & \textbf{Con.} & \textbf{Avg.} \\
\midrule
\multicolumn{6}{l}{\textit{LLMs}} \\
GPT-4
                    & 98.40 & 83.10 & 88.90 & 94.10 & 91.12 \\
GPT-4o
                    & 98.30 & 84.50 & 87.20 & 94.30 & 91.08 \\
GPT-4o-mini
                    & 97.30 & 74.30 & 90.30 & 94.40 & 89.08 \\
Gemini-1.5-Pro      & 97.90 & 80.30 & 92.00 & 95.40 & \textbf{91.40} \\
Gemini-2.0-Flash    & 98.20 & 85.50 & 85.40 & 93.40 & 90.62 \\
Claude-3.7-Sonnet
                    & 97.10 & 75.90 & 93.00 & 95.90 & 90.47 \\
DeepSeek-v3
                    & 99.10 & 85.20 & 87.40 & 93.30 & 91.25 \\
Mistral Large
                    & 98.10 & 81.30 & 88.70 & 94.60 & 90.68 \\
\midrule
\multicolumn{6}{l}{\textit{Sentence Embedding Models}} \\
E5-base-v2          & 85.00     & 58.70     & 68.30     & 52.90     & \textbf{66.22} \\
GTE-base            & 75.70     & 51.00     & 76.80     & 57.20     & 65.18 \\
EmbeddingGemma      & 77.10     & 51.20     & 65.70     & 54.70     & 62.17 \\
\midrule
\multicolumn{6}{l}{\textit{BERT-based}} \\
SimCSE (SNLI+MNLI)  & 78.50 & 41.00 & 77.40 & 67.50 & 66.10 \\
SimCSE (INLI)       & 89.80 & 67.60 & 65.70 & 83.90 & 76.75 \\
ImpScore (INLI)     & 59.20 & 26.30 & 75.30 & 81.50 & 60.58 \\
DualCSE-Cross (ours)& 86.80 & 64.30 & 72.40 & 87.50 & 77.75 \\
DualCSE-Bi (ours)   & 91.30 & 63.30 & 73.60 & 85.10 & \textbf{78.32} \\
\midrule
\multicolumn{6}{l}{\textit{RoBERTa-based}} \\
SimCSE (SNLI+MNLI)  & 79.80 & 49.00 & 74.30 & 67.60 & 67.68 \\
SimCSE (INLI)       & 90.60 & 69.10 & 66.90 & 91.00 & 79.40 \\
ImpScore (INLI)     & 81.60 & 56.80 & 47.70 & 61.60 & 61.92 \\
DualCSE-Cross (ours)& 90.20 & 73.40 & 68.40 & 88.70 & 80.18 \\
DualCSE-Bi (ours)   & 91.90 & 69.90 & 72.10 & 87.60 & \textbf{80.38} \\
\bottomrule
\end{tabular}
\caption{Full results of the RTE task (accuracy \%)}
\label{tab:rte_results_full}
\end{table}

\section{Full Results of EIS}
\label{app:full_results_of_eis}
The full results of the EIS task are shown in Table~\ref{tab:eis_results_full}. 
It is noteworthy that DualCSE-Cross outperforms ImpScore (INLI) when BERT is used as the base encoder, whereas they are comparable when RoBERTa is used. 

\begin{table}[!ht]
\centering
\small
\begin{tabular}{lcc}
\toprule
\textbf{Model}          & \textbf{INLI} & \textbf{\citet{wang2025impscore}} \\
\midrule
\textsc{Length}         & 99.90         & 73.37 \\
ImpScore (original)     & 80.55         & \textbf{95.20} \\
\midrule
\multicolumn{3}{l}{\textit{BERT-based}} \\
ImpScore (INLI)         & 99.97         & 76.91 \\
DualCSE-Cross (ours)    & \textbf{100}  & 80.46 \\
DualCSE-Bi (ours)       & 99.97         & 79.88 \\
\midrule
\multicolumn{3}{l}{\textit{RoBERTa-based}} \\
ImpScore (INLI)         & 99.97         & 81.56 \\
DualCSE-Cross (ours)    & 99.97         & 79.31 \\
DualCSE-Bi (ours)       & \textbf{100}  & 77.48 \\
\bottomrule
\end{tabular}
\caption{Full results of the EIS task (accuracy \%)}
\label{tab:eis_results_full}
\end{table}

\section{Ablation Details}
\label{app:ablation_results}
The detail description of ablation experiments are follows. 
\paragraph{w/o contradiction}
We remove $v(\mathbf{r}_i, \mathbf{r}^-_j)$ and $v(\mathbf{u}_i, \mathbf{r}^-_j)$ in the denominator and $- \log{\frac{v(\mathbf{r}^-_i, \mathbf{u}^-_{i})}{\sum^{N}_{j=1}v(\mathbf{r}^-_i, \mathbf{u}^-_j)}}$ from Equation~(\ref{eq:loss}). 
The entire formula is as follows: 
\begin{equation}
    \small
    \begin{array}[b]{@{}l@{\;}l@{}}
        l_i = 
        & \displaystyle
        -\log{\frac{v(\mathbf{r}_i, \mathbf{r}^+_{i1})}{\sum^{N}_{j=1}(v(\mathbf{r}_i, \mathbf{r}^+_{j1}) + v(\mathbf{r}_i, \mathbf{u}_j))}} \\
        & \displaystyle
        - \log{\frac{v(\mathbf{u}_i, \mathbf{r}^+_{i2})}{\sum^{N}_{j=1}(v(\mathbf{u}_i, \mathbf{r}^+_{j2}) + v(\mathbf{u}_i, \mathbf{r}_j))}} \\
        & \displaystyle
        - \log{\frac{v(\mathbf{r}^+_{i1}, \mathbf{u}^+_{i1})}{\sum^{N}_{j=1}v(\mathbf{r}^+_{i1}, \mathbf{u}^+_{j1})}} - \log{\frac{v(\mathbf{r}^+_{i2}, \mathbf{u}^+_{i2})}{\sum^{N}_{j=1}v(\mathbf{r}^+_{i2}, \mathbf{u}^+_{j2})}}. 
    \end{array}
\end{equation}
\paragraph{w/o intra-sentence}
We remove $v(\mathbf{r}_i, \mathbf{u}_j)$ and $v(\mathbf{u}_i, \mathbf{r}_j)$ in the denominator and $- \log{\frac{v(\mathbf{r}^+_{i1}, \mathbf{u}^+_{i1})}{\sum^{N}_{j=1}v(\mathbf{r}^+_{i1}, \mathbf{u}^+_{j1})}}$,  $ - \log{\frac{v(\mathbf{r}^+_{i2}, \mathbf{u}^+_{i2})}{\sum^{N}_{j=1}v(\mathbf{r}^+_{i2}, \mathbf{u}^+_{j2})}}$ and $- \log{\frac{v(\mathbf{r}^-_i, \mathbf{u}^-_{i})}{\sum^{N}_{j=1}v(\mathbf{r}^-_i, \mathbf{u}^-_j)}}$ from Equation~(\ref{eq:loss}). 
The entire formula is as follows: 
\begin{equation}
    \small
    \begin{array}[b]{@{}l@{\;}l@{}}
        l_i =
        & \displaystyle
        -\log{\frac{v(\mathbf{r}_i, \mathbf{r}^+_{i1})}{\sum^{N}_{j=1}(v(\mathbf{r}_i, \mathbf{r}^+_{j1}) + v(\mathbf{r}_i, \mathbf{r}^-_j))}} \\
        & \displaystyle
        - \log{\frac{v(\mathbf{u}_i, \mathbf{r}^+_{i2})}{\sum^{N}_{j=1}(v(\mathbf{u}_i, \mathbf{r}^+_{j2}) + v(\mathbf{u}_i, \mathbf{r}^-_j))}}. 
    \end{array}
\end{equation}

\paragraph{w/o contradiction \& intra-sentence}
The formula of the loss function is as follows: 
\begin{equation}
    \small
    \begin{array}[b]{@{}l@{\;}l@{}}
        l_i =
        & \displaystyle
        -\log{\frac{v(\mathbf{r}_i, \mathbf{r}^+_{i1})}{\sum^{N}_{j=1}(v(\mathbf{r}_i, \mathbf{r}^+_{j1}))}} \\
        & \displaystyle
        - \log{\frac{v(\mathbf{u}_i, \mathbf{r}^+_{i2})}{\sum^{N}_{j=1}(v(\mathbf{u}_i, \mathbf{r}^+_{j2}))}}. 
    \end{array}
\end{equation}

\begin{table*}[t]
    \centering
    \small
    \begin{subtable}{1.0\linewidth}
    \begin{tabular}{@{}p{0.485\linewidth}|p{0.485\linewidth}@{}}
        \toprule
        \multicolumn{2}{@{}p{0.99\linewidth}@{}}{\textbf{Query}: Joseph wants to know about Fred's food preferences. Joseph says, ``Would you be into eating at a diner with burgers?'' Fred responds, ``I want to get a salad.''} \\
        \midrule
        \textbf{Explicit semantic}: Fred wants to get a salad. & \textbf{Implicit semantic}: It's unlikely that Fred wants to eat burgers at a diner. \\
        \midrule
        \#1 Terrie prefers salads (to food served at fast food restaurants). & \#1 Cookies are something that Fred enjoys. \\
        \#2 Hannah will travel up to five miles, but only for a salad. & \#2 Fredrick enjoys spicy food, but only if he has milk to cool his mouth. \\
        \#3 Marcus says, ``That'd be great,'' in response to Normand's suggestion of a vegetarian restaurant. & \#3 Freddie believes that pizza would be a good food choice. \\
        \bottomrule
    \end{tabular}
    \caption{Example \#1}
    \bigskip
    \end{subtable}
    \begin{subtable}{1.0\linewidth}
    \begin{tabular}{@{}p{0.485\linewidth}|p{0.485\linewidth}@{}}
        \toprule
        \multicolumn{2}{@{}p{0.99\linewidth}@{}}{\textbf{Query}: Pete says, ``That chocolate cake looks delicious. Aren't you going to have some with me?'' Connie responds, ``I am allergic to chocolate.''} \\
        \midrule
        \textbf{Explicit semantic}: Connie claims to have an allergy to chocolate. & \textbf{Implicit semantic}: Connie will not join Pete in eating chocolate cake. \\
        \midrule
        \#1 Francisco says, ``It is too cold,'' when Vickie asks if he wants to go swimming. & \#1 Francis cannot eat certain foods. \\
        \#2 Christie says that she and Peter are completely different. & \#2 Elva won't eat any cake. \\
        \#3 Katie doesn't like the thing that Cristina is talking about. & \#3 Carmen prefers not to eat at the restaurant. \\
        \bottomrule
    \end{tabular}
    \caption{Example \#2}
    \bigskip
    \end{subtable}
    \begin{subtable}{1.0\linewidth}
    \begin{tabular}{@{}p{0.485\linewidth}|p{0.485\linewidth}@{}}
        \toprule
        \multicolumn{2}{@{}p{0.99\linewidth}@{}}{\textbf{Query}: Phoebe says, ``Do you like my new outfit?'' Rolland responds, ``You shouldn't be allowed to buy clothes.''} \\
        \midrule
        \textbf{Explicit semantic}: Rolland believes Phoebe should be prevented from purchasing clothes. & \textbf{Implicit semantic}: Rolland really hates Phoebe's new outfit. \\
        \midrule
        \#1 Rosendo claims he did not order the code red. & \#1 The item is too big for her, so it won't be suitable. \\
        \#2 Rolland does not have any children. & \#2 Alphonso will not be going shopping. \\
        \#3 Rosendo doesn't think listening to local indie artists is cool. & \#3 Lois has no desire to go to the mall. \\
        \bottomrule
    \end{tabular}
    \caption{Example \#3}
    \end{subtable}

    \caption{Several examples of a simple retrieval experiment}
    \label{tab:retrieval_example2}
\end{table*}

The full results of the ablation experiments are shown in Table~\ref{tab:ablation_results_full}. 

\begin{table}[!ht]
    \centering
    \small
    \begin{tabular}{lcc}
        \toprule
        \textbf{Loss function}                  & \textbf{RTE}      & \textbf{EIS} \\
        \midrule
        DualCSE-Cross-BERT                      & \textbf{77.75}    & \textbf{100} \\
        ~~w/o contradiction                     & 64.13             & 99.90 \\
        ~~w/o intra sentence                    & 77.50             & 47.13 \\
        ~~w/o contradiction \& intra sentence   & 64.38             & 31.83 \\
        \midrule
        DualCSE-Bi-BERT                         & \textbf{78.32}    & 99.97 \\
        ~~w/o contradiction                     & 65.97             & \textbf{100} \\
        ~~w/o intra sentence                    & 77.30             & 63.42 \\
        ~~w/o contradiction \& intra sentence   & 65.47             & 81.35 \\
        \midrule
        DualCSE-Cross-RoBERTa                   & \textbf{80.18}    & \textbf{99.97} \\
        ~~w/o contradiction                     & 64.57             & 99.88 \\
        ~~w/o intra sentence                    & 80.10             & 92.25 \\
        ~~w/o contradiction \& intra sentence   & 64.68             & 32.75 \\
        \midrule
        DualCSE-Bi-RoBERTa                      & 80.38             & \textbf{100} \\
        ~~w/o contradiction                     & 66.13             & 99.95 \\
        ~~w/o intra sentence                    & \textbf{80.57}    & 60.35 \\
        ~~w/o contradiction \& intra sentence   & 65.07             & 76.15 \\
        \bottomrule
    \end{tabular}
    \caption{Full results of ablation experiments}
    \label{tab:ablation_results_full}
\end{table}

\section{Examples of Retrieval Experiment}
\label{app:retrieval_examples}
Several examples of the retrieval experiment are shown in Table~\ref{tab:retrieval_example2}. 

\end{document}